\documentclass{article}
\usepackage{iclr2026_conference,times}

\usepackage{amsmath,amsfonts,bm}

\def\eqref#1{equation~\ref{#1}}

\def\1{\bm{1}}

\DeclareMathAlphabet{\mathsfit}{\encodingdefault}{\sfdefault}{m}{sl}
\SetMathAlphabet{\mathsfit}{bold}{\encodingdefault}{\sfdefault}{bx}{n}

\usepackage{hyperref}
\usepackage{xurl}  
\usepackage[capitalise]{cleveref}

\crefformat{section}{#2\S#1#3}
\crefname{figure}{Fig.}{Figs.}
\crefname{table}{Tab.}{Tabs.}
\usepackage{graphicx}   
\usepackage{hyperref}       
\usepackage{multirow}
\usepackage{multicol}
\usepackage{array}
\usepackage{subcaption}
\usepackage{booktabs}
\usepackage{xcolor}
\usepackage{xspace}
\usepackage{listings}   
\usepackage{wrapfig}
\usepackage[frozencache,cachedir=minted-cache]{minted}
\usepackage{booktabs}
\usepackage[table]{xcolor}
\usepackage[most]{tcolorbox}
\usepackage{wrapfig}

\tcbuselibrary{listings, skins}
\usepackage{listings}
\definecolor{col_overview_a}{RGB}{59,126,161}

\newcommand*\circlea[1]{\tikz[baseline=(char.base)]{%
		\node[white,shape=circle,fill=col_overview_a,draw,inner sep=1pt] (char) {\color{white}\fontsize{8pt}{8pt}\selectfont\sffamily #1};}}

\definecolor{BerkeleyBlue}{HTML}{003262}
\definecolor{FoundersRock}{HTML}{3B7EA1}
\definecolor{CodeBg}{HTML}{F4F6F7} 

\title{
SkyRL-Agent: Efficient RL Training for Multi-turn LLM Agent}

\author{Shiyi Cao$^{\S*}$, Dacheng Li$^{\S}$\thanks{Co-lead the project. $^\dagger$ Main Contributors.}\;\,, Fangzhou Zhao$^{\S^\dagger}$, Shuo Yuan$^{\S^\dagger}$, Sumanth R Hegde$^{\P^\dagger}$, \\
\textbf{Connor Chen$^{\S^\dagger}$, Charlie Ruan$^{\S^\dagger}$, Tyler Griggs$^{\S}$, Shu Liu$^{\S}$, Eric Tang$^{\P}$, Richard Liaw$^{\P}$,} \\ 
\textbf{Philipp Moritz$^{\P}$, Matei Zaharia$^{\S}$, Joseph E. Gonzalez$^{\S}$, Ion Stoica$^{\S}$}\\
NovaSky AI, $^{\S}$UC Berkeley, $^{\P}$Anyscale
\\[3pt]
\raisebox{-1.5pt}{\includegraphics[height=1.05em]{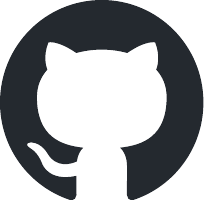}}~\texttt{GitHub:}
\url{https://github.com/NovaSky-AI/SkyRL}
\\
\raisebox{-2.5pt}{\includegraphics[height=1.05em]{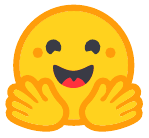}}~\texttt{HuggingFace:}
\url{https://huggingface.co/NovaSky-AI/SA-SWE-32B}
}

\newcommand{\frameworkname}{\textsc{SkyRL-Agent}\xspace}
\newcommand{\model}{SA-SWE-32B\xspace}

\iclrfinalcopy 
\begin{document}

\maketitle

\begin{abstract}

We introduce \frameworkname, a framework for efficient, multi-turn, long-horizon agent training and evaluation. It provides efficient asynchronous dispatching, lightweight tool integration, and flexible backend interoperability, enabling seamless use with existing RL frameworks such as SkyRL-train, VeRL, and Tinker.

Using \frameworkname, we train \model, a SWE agent trained from Qwen3-32B (24.4\% Pass@1) purely with RL.
We introduce two key components: an optimized asynchronous pipeline dispatcher that achieves a 1.55$\times$ speedup over naive asynchronous batching, and a tool-enhanced training recipe leveraging an AST-based search tool to facilitate code navigation, boost rollout Pass@K, and improve training efficiency.
Together, these optimizations enable \model to reach 39.4\% Pass@1 on SWE-Bench Verified\footnote{Evaluated using a simplified ReAct loop with only the file-editor and bash tools, under 40k context length and 100 max steps.} with more than 2$\times$ cost reduction than prior models reaching similar performance. Despite being trained solely on SWE tasks, \model generalizes effectively to other agentic tasks, including Terminal-Bench, BrowseComp-Plus, and WebArena. We further demonstrate \frameworkname’s extensibility through case studies on deep research, computer use, and memory agents, each trained using a different training backend.

\end{abstract}


\begin{figure}[ht]
\vspace{-2mm}
    \centering
    \begin{subfigure}[b]{0.54\linewidth}
        \centering
        \includegraphics[width=\linewidth]{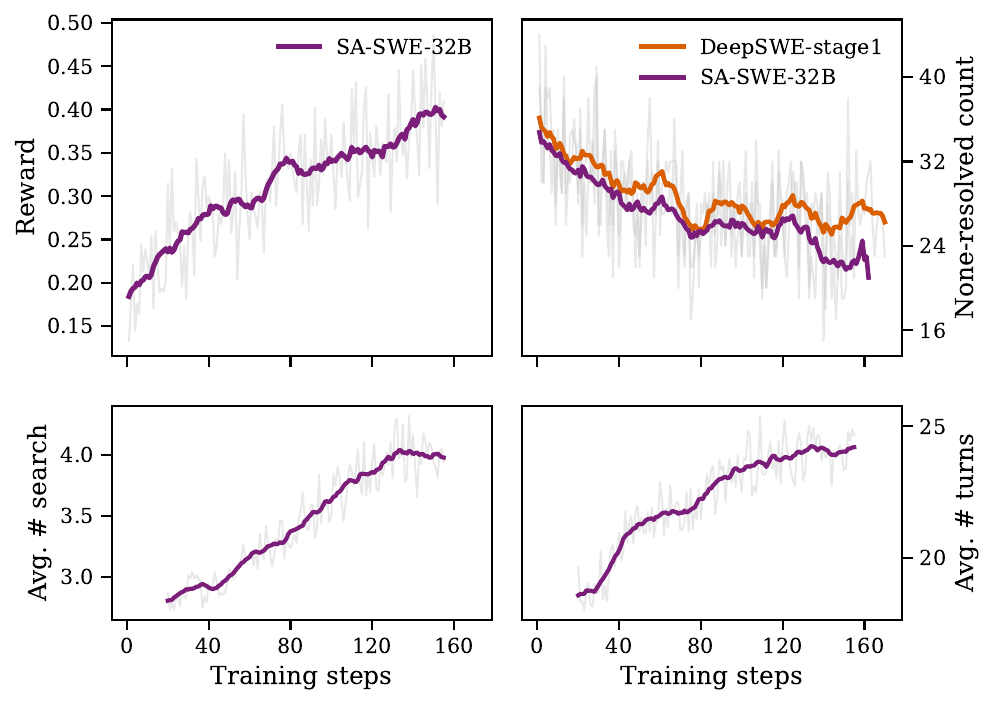}
        \caption{Training metrics for SA-SWE-32B and baselines. We use the checkpoint at step 125 for final evaluation. Curves for DeepSWE are taken from their wandb logs.}
        \label{fig:swe_reward}
    \end{subfigure}
    \hspace{2mm}
    \begin{subfigure}[b]{0.4\linewidth}
        \centering
        \includegraphics[width=\linewidth]{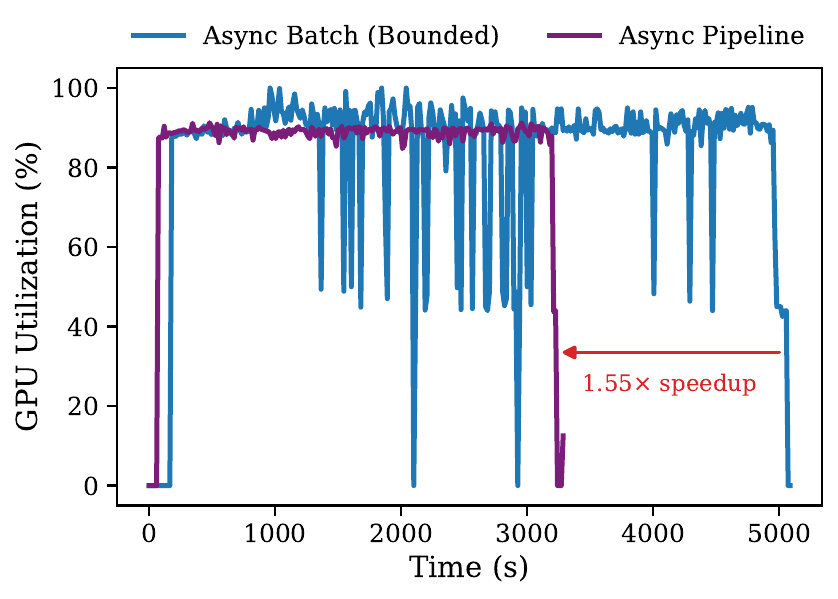}
        \caption{GPU utilization during generation for \textbf{Async Pipeline} and \textbf{Async Batch (Bounded)}. Evaluated under batch size 64 and 8 rollouts (512 total) with 2$\times$8 H100 GPUs for \model training.}
        \label{fig:dispatch_bench}
    \end{subfigure}
    \caption{Training dynamics and system performance for SA-SWE-32B. (a) compares training metrics across ablations; (b) shows GPU utilization comparison between asynchronous dispatching strategies.}
    \label{fig:train_sys_combined}
\end{figure}

\section{Introduction}

Recent progress in post-training techniques, especially \emph{reinforcement learning from verifiable rewards (RLVR)}~\citep{guo2025deepseek,hu2025open,wen2025reinforcement}, has spurred a growing effort to move beyond single-turn language model optimization toward the training of \emph{multi-turn, long-horizon, tool-augmented agents}~\citep{jin2025search,chen2017retool,gao2025beyond,li2025webthinker,li2025torl,luo2025agent,cao2025skyrl}. These agents are capable of performing complex, multi-step reasoning and acting across diverse environments such as code repositories, browsers, or operating systems. As tasks become increasingly interactive and open-ended, training such agents efficiently and reliably has emerged as a major challenge for both the reinforcement learning (RL) and systems communities.

Recent RL training frameworks such as SkyRL-train~\citep{griggs2025evolving}, AReaL~\citep{fu2025areal}, VeRL~\citep{sheng2025hybridflow}, and SLIME~\citep{slime_github} focus on improving the systems efficiency of large-scale reinforcement learning for language models. These frameworks manage device placement and data flow across heterogeneous clusters, coordinate efficient weight synchronization between inference and training engines, and employ diverse parallelization strategies (e.g., tensor, data, and pipeline parallelism) together with asynchronous or hybrid execution plans to maximize hardware utilization. 
Tinker~\citep{tinker} builds on this ecosystem by offering modular APIs for fine-tuning and sampling, allowing users to experiment with customized training algorithms and data pipelines while abstracting away the complexity of distributed execution.

Unfortunately, what is still missing is a modular and performant \emph{agentic rollout orchestration layer} that can generate, schedule, and evaluate multi-turn agent trajectories at scale. In practice, researchers and practitioners are forced to build ad-hoc solutions for coordinating asynchronous rollouts, managing environment and agent states, and integrating to existing training frameworks. Such fragmented setups result in inefficiencies, brittle execution, and debugging challenges, especially given the inherent instability and fragility of RL training.


Specifically, the framework should satisfy the following key properties to enable efficient, scalable, and extensible agent training: 
\begin{enumerate}

\item \textbf{Flexible tool and task integration.}
LLM-based agents rely on a wide range of tools to complete tasks, including stateless utilities (e.g, python interpreters), environment-modifying tools (e.g., file editors), and agent-state-modifying operations (e.g., summarization and history truncation), each with distinct runtime requirements.  
The framework should allow new tasks and tools to be integrated easily with minimum modifications to the main agent loop.

\item \textbf{Efficient rollout scheduling.}
To achieve high hardware utilization and minimize latency, the framework should support fine-grained scheduling rather than treating each rollout as a monolithic job.
\emph{Intra-rollout scheduling} breaks a rollout into stages (e.g., initialization, LLM generation, reward computation) and schedules these stages independently, enabling their heterogeneous CPU- and GPU-bound operations to overlap with stages from other rollouts.
\emph{Inter-rollout scheduling} determines the global dispatch order and prioritization of rollouts so that workloads across CPU and GPU remain balanced over time, preventing hardware idling and reducing overall makespan.
Together, these mechanisms improve heterogeneous resource utilization and overall generation throughput.

\item \textbf{Training-backend agnostic.}  
Developers should be able to specify the agent design, tool execution logic, and training backend independently. Decoupling these components allows a single agent implementation to run unmodified across different frameworks, to flexibly leverage their unique capabilities and seamlessly transition between local deployments and API-based services.

\end{enumerate}

\frameworkname builds on these observations by introducing an efficient execution framework for heterogeneous agent tasks. It consists of three key components:
(1) a \emph{tool-centric task interface} that supports dynamic registration of user-defined tools, instruction builders, and verifiers for different tasks, enabling seamless integration of new tasks and tools with minimal code changes; (2) a \emph{fine-grained asynchronous dispatcher abstraction} that provides a unified interface for designing scheduling policies for a batch of multi-stage rollout jobs; and
(3) a \emph{backend bridge} that connects to RL training systems such as Tinker, VeRL, and SkyRL-train for advantage estimation and policy optimization.

To demonstrate the effectiveness of \frameworkname, we use it to train \model, a software-engineering agent derived from Qwen3-32B (24.4\% Pass@1), using pure RL on 4.5K instances from R2E-Gym~\citep{jain2025r2e}.
\model achieves 39.4\% Pass@1 on SWE-Bench Verified~\citep{jimenez2024swebench}, matching the performance of state-of-the-art models of similar scale while reducing the total training cost by more than \textbf{2$\times$}.
This efficiency gain stems from two key contributions: an \emph{asynchronous pipeline dispatcher} that delivers a \textbf{1.55$\times$} speedup over naive asynchronous batching by overlapping CPU- and GPU-bound operations (\cref{sec:design}), and a \emph{tool-enhanced training recipe} leveraging an AST-based search tool that facilitates code navigation, leading to higher Pass@K for rollouts and improved sample efficiency (\cref{sec:swe_recipe}).
Although trained exclusively on software-engineering tasks, \model generalizes effectively to other agentic benchmarks (\cref{sec:swe_eval}) such as Terminal-Bench~\citep{tbench_2025}, BrowseComp-Plus~\citep{chen2025browsecomp}, and WebArena~\citep{zhou2023webarena}.

Beyond the SWE agent, we further demonstrate the versatility of \frameworkname in \cref{sec:other_agents} by training several additional agents in smaller scale, including the \textit{Deep Research Agent}, the \textit{Computer Use Agent}, and the \textit{Memory Agent}, each connected to distinct RL backends, showcasing the framework’s flexibility and interoperability.

\section{Background}






\begin{table*}[ht]
\centering
\small
\renewcommand{\arraystretch}{1.1}
\setlength{\tabcolsep}{3pt}
\caption{
\textbf{Comparison of \frameworkname with existing agent training frameworks.}
\frameworkname offers unified tool interface, efficient rollout scheduling, backend portability, runtime scaling utilities, and dynamic trajectory construction for scalable agent training.
}
\resizebox{0.98\textwidth}{!}{
\begin{tabular}{lccccc}
\toprule
\textbf{Framework} &
\textbf{Backend} &
\textbf{Agent Execution} &
\textbf{Interface} &
\textbf{Runtime Scaling} &
\textbf{Trajectory} \\
\midrule
\textbf{VeRL-Tool}~\citep{jiang2025verltool} &
VeRL &
Data-parallel &
Tool &
Ray &
Mask \\

\textbf{rLLM}~\citep{rllm2025} &
VeRL &
Data-parallel &
Gym &
K8s &
Mask / Transition \\

\textbf{GEM}~\citep{liu2025gem} &
Multi &
Data-parallel &
Gym &
-- &
Transition \\

\textbf{Agent-Lightning}~\citep{luo2025agent} &
Multi &
Data-parallel &
-- &
-- &
Transition \\

\midrule
\textbf{SkyRL-Agent (Ours)} &
\textbf{Multi} &
\textbf{Data + Pipeline (extensible)} &
\textbf{Tool} &
\textbf{Ray/K8s} &
\textbf{Mask / Transition} \\
\bottomrule
\end{tabular}
}
\label{tab:framework_comparison}
\end{table*}

\paragraph{Multi-Turn LLM Agent Training.}
Reinforcement learning (RL) has shown strong potential for improving \emph{single-turn reasoning tasks} such as mathematics, logic, and coding~\citep{shao2024deepseekmath,guo2025deepseek,hu2025open,Polaris2025,yu2025dapo}. 
In these settings, the model generates a complete response to a single prompt and receives a scalar reward reflecting its correctness or quality. 
Building on these successes, researchers have further explored \emph{tool-integrated reasoning}, where the model interacts with external tools such as search engines, calculators, or code executors~\citep{li2025torl,jin2025search,chen2017retool} to enhance its reasoning abilities, showing significant improvement on related reasoning benchmarks. 
Since tool responses (e.g., retrieved documents or computed results) are interleaved with the model’s own reasoning, most approaches employ a \emph{mask-based loss construction}, where non-model-generated tokens are masked during optimization to ensure that only generated segments contribute to the policy gradient. 

Recent works have also applied RL to \emph{long-horizon, multi-turn agent training}~\citep{cao2025skyrl,deepswe2025,gao2025beyond,lu2025arpo}, enabling optimization of agent scaffolds such as ReAct~\citep{yao2022react}, CodeAct~\citep{wang2024executable}, and MemGPT~\citep{packer2023memgpt} beyond supervised learning~\citep{pan2024training,yang2025swe}. 
A multi-turn agent can be formulated as a partially observable Markov decision process (POMDP)~\citep{kaelbling1998planning}, where at each step \(t\), the agent observes a context \(o_t\), generates an action \(a_t\) from a policy \(\pi_\theta(a_t|o_t)\), and receives a scalar reward \(r_t\) from the environment. 
The objective is to maximize the expected cumulative return
\[
J(\theta) = \mathbb{E}_{\pi_\theta}\Big[\sum_{t=1}^{T} r_t \Big],
\]
where \(T\) is the number of agent turns per task.

In multi-turn settings, agents often modify their context between steps through summarization, truncation, or selective incorporation of retrieved information, which makes the overall interaction history non-stationary. 
Such dynamic context management breaks the assumption of a single continuous text sequence, making the conventional \emph{mask-based method} difficult to generalize beyond short, tool-integrated reasoning tasks. 

Building on the POMDP formulation, recent work instead adopts a more general \emph{transition-based} data representation~\citep{rllm2025,luo2025agent}, where each model invocation and its resulting feedback are represented as a tuple \((o_t, a_t, r_t)\). 
This representation supports flexible reward assignment and decouples learning from the details of agent execution and context management. It also naturally applies to multi-agent or sub-agent systems, where different policies interact within a shared environment. 



\paragraph{Training Frameworks for LLM Agents.}
 Existing agent training frameworks such as VeRL-Tool~\citep{jiang2025verltool}, rLLM~\citep{rllm2025}, GEM~\citep{liu2025gem}, and Agent-Lightning~\citep{luo2025agent} have advanced large-scale RL training for LLM agents, yet remain limited by naive asynchronous batching (i.e., data-parallel) execution. \frameworkname supports flexible rollout scheduling in multiple dimensions (e.g., data-parallel + pipeline-parallel).
 Beyond execution efficiency, \frameworkname unifies stateless tools (e.g., python interpreter), environment-modifying actions (e.g., file editor), and agent-state-modifying operations (e.g., summarization) within a single tool, unlike Gym-centric loops~\citep{rllm2025, liu2025gem}, which manage agent state outside \texttt{env.step} with ad hoc implementations.
As summarized in \cref{tab:framework_comparison}, Agent-Lightning serves as a middle layer between training backends and workflow frameworks such as LangChain~\citep{LangChain} and AutoGen~\citep{AutoGen}, but lacks native support for user-defined tools and new task integration. rLLM and VeRL-Tool provide stronger extensibility, but they are tied to the VeRL~\citep{sheng2025hybridflow} backend, and VeRL-Tool’s mask-based trajectory construction limits its applicability to settings such as multi-agent systems or memory agents.

\section{Framework Architecture}

\begin{figure}[ht!]
    \centering
    \includegraphics[width=0.9\linewidth]{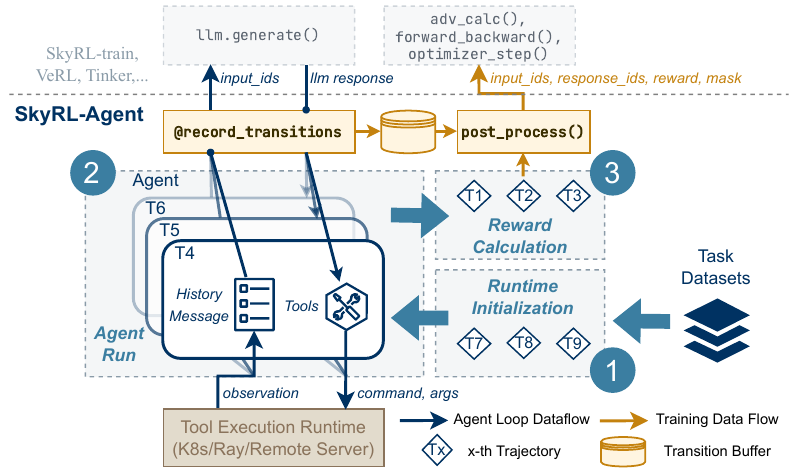} 
    \caption{Overview Architecture of \frameworkname. The framework decomposes each rollout into three stages: (1) \emph{runtime initialization} for tool execution runtime setup, (2) \emph{agent run} where the agent performs actions through the tool interface, and (3) \emph{reward calculation} for outcome evaluation. 
During execution, the inputs and outputs of LLM calls are recorded as transitions and stored in a buffer, while \texttt{post\_process} aggregates these transitions together with their rewards into formatted data compatible with multiple RL training backends such as SkyRL-train, VeRL, and Tinker. 
The dispatcher schedules jobs across the three stages according to predefined policies.}
    \label{fig:workflow}
\end{figure}

\frameworkname is designed as a modular framework for training and evaluating tool-use agents at scale. 
The system architecture is illustrated in \cref{fig:workflow}, consisting of three main components: a \emph{tool-centric agent loop} (\cref{sec:tool_interface}) for flexible tool and task integration, a \emph{fine-grained dispatcher} (\cref{sec:design}) for heterogeneous scheduling, and a \emph{backend bridge} (\cref{sec:backend}) for seamless connection with RL training systems.

\subsection{Tool-centric agent loop.}\label{sec:tool_interface}

In \frameworkname, agents act purely via
OpenAI-style function calls. Each tool implements its own execution logic and specifies its runtime. Existing Gym-style environments~\citep{liu2025gem,jain2025r2e, OSWorld}, can be integrated by wrapping their \texttt{step()} as a tool, for example, an implementation of tools for the computer use agent is shown in~\cref{lst:osworld}.

This design yields several practical benefits:
(i) \emph{unified management of agent and environment states}: unlike Gym-centric agent loops~\citep{rllm2025, liu2025gem}, where agent state is managed outside \texttt{env.step} through ad hoc code, \frameworkname brings it under the same tool abstraction as other actions, making agent-state-modifying operations such as context management modular and learnable.
(ii) \emph{convenient multi‑task training}: different datasets bind to different tool sets and verifiers without modifying the agent loop, allowing the runner to multiplex tasks within a single training job; and (iii) \emph{minimal‑change task integration}: adding a new task reduces to providing a small set of tool implementations together with task‑specific instruction builder and verifiers, leaving agent code unchanged and without touching other parts of the system.


\begin{listing}[ht]
\begin{minted}[
  fontsize=\small,
  breaklines,
  breakanywhere,
  baselinestretch=0.95,
  bgcolor=CodeBg,  % <--- background color here
  fontfamily=zi4  % uses Inconsolata if available
]{python}
@register_tool("osworld_action")
class OSWorldActionTool(BaseTool):
    name = "osworld_action"
    description = (
        "Execute desktop automation actions using pyautogui. "
        "Provide Python code such as 'pyautogui.click(500, 300)' "
        "or 'pyautogui.typewrite(\"Hello World\")'."
    )
    parameters = {
        "type": "object",
        "properties": {"code": {"type": "string"}},
        "required": ["code"]
    }

    def call(self, params: Union[str, dict], runtime=None, **kwargs) -> str:
        ...
        code = params.get("code", "").strip()
        obs, reward, done, info = runtime.step(code, TIMEOUT)
        ...
\end{minted}
\caption{Example definition for desktop tools in the Computer Use agent.}
\label{lst:osworld}
\end{listing}

\subsection{Fine‑grained heterogeneous scheduling via the dispatcher.}\label{sec:design}
Multi‑turn RL rollouts comprise operations with disparate costs and device affinities. As illustrated in~\cref{fig:workflow}, \frameworkname decomposes each trajectory into stage jobs: \circlea{1}~\emph{runtime initialization}, \protect\circlea{2}~\emph{agent run}, and \protect\circlea{3}~\emph{reward calculation}. The dispatcher maintains bounded queues per stage and routes jobs according to predefined policies, balancing workloads on heterogeneous devices to improve overall resource utilization.

\begin{figure}[ht]
    \centering
    \includegraphics[width=\linewidth]{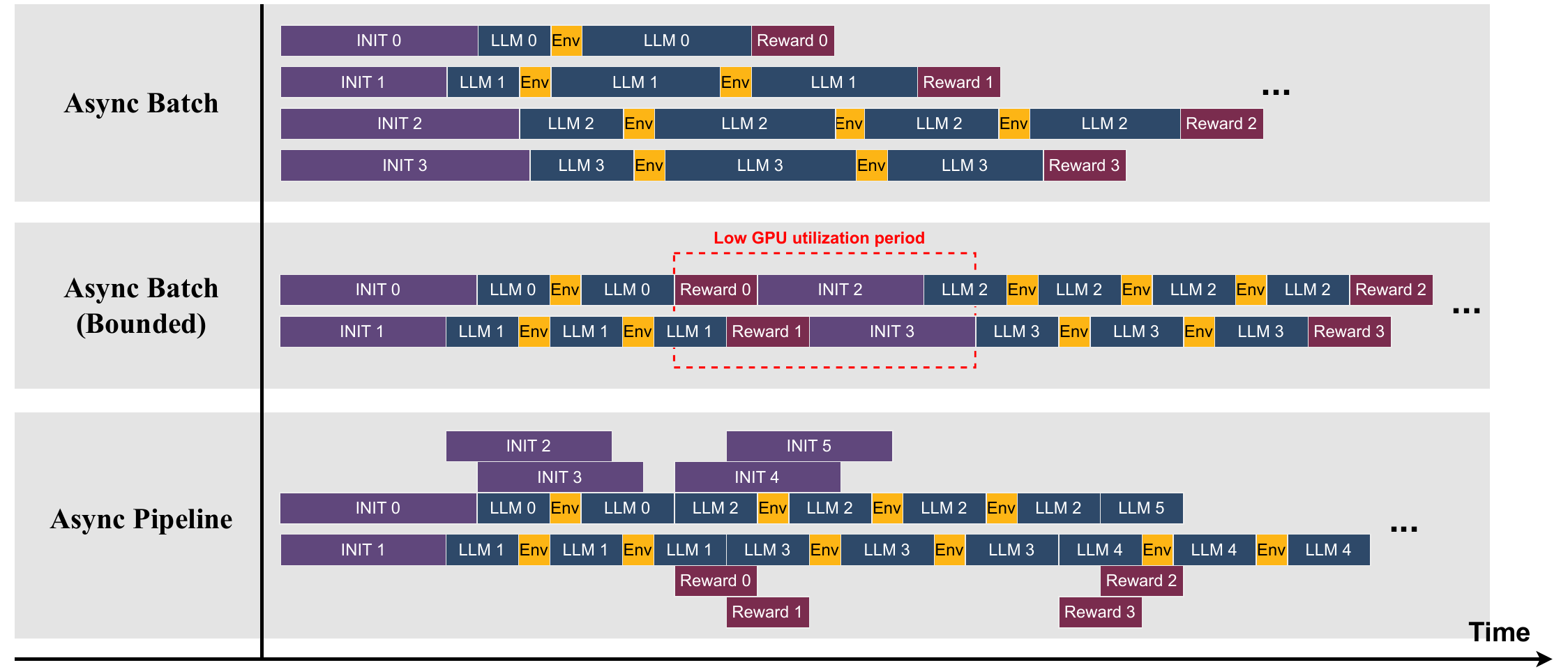} 
    \caption{Examples of Supported Dispatching Methods.
    Async Batch is normally used for reasoning tasks where runtime initialization and reward computation are lightweight. Async Batch (Bounded) schedules trajectories sequentially with capped concurrency, leading to unbalanced GPU utilization across stages, but remains effective when runtime reset is inexpensive, such as in computer-use tasks. Async Pipeline overlaps the three stages to maintain high GPU utilization, suitable for tasks with expensive runtime or reward stages.}
    \label{fig:dispatcher}
\end{figure}

We provide several dispatching strategies under a unified interface, allowing practitioners to conveniently select the policy that best matches the characteristics of their tasks, as illustrated in \cref{fig:dispatcher}.
\emph{Async Batch} launches all trajectories concurrently and is effective when both runtime initialization and reward calculation are lightweight, such as in search-integrated reasoning tasks~\citep{jin2025search}.
\emph{Async Batch (Bounded)} limits concurrency to a configurable pool size, which is suitable for environments where concurrency must be restricted to prevent overloading the runtime or where persistent resources (e.g., long-lived virtual machines in computer-use tasks) can be efficiently reused across rollouts.
\emph{Async Pipeline} employs three bounded queues of different, configurable sizes to overlap CPU-bound stages with GPU-bound inference, preventing CPU overhead from throttling the agent loop and overload of runtime services. Existing agent training frameworks typically adopt the asynchronous batching approaches.


In our SWE agent training (\cref{sec:swe_recipe}), the \emph{Async Pipeline} dispatching strategy achieves approximately a \textbf{1.55$\times$} speedup over the naive \emph{Async Batch (Bounded)} approach, as shown in \cref{fig:dispatch_bench}. With the async pipeline method, GPU utilization remains stable at around 90\% throughout the generation stage. In contrast, the async batch strategy exhibits large fluctuations and frequent drops in utilization, revealing bubbles on the GPU side caused by CPU-bound operations such as runtime initialization and reward computation. During these intervals, the GPU stays idle while waiting for CPU-side preparation to complete. The asynchronous pipeline approach eliminates these idle periods by overlapping CPU- and GPU-bound stages, keeping the GPU consistently active and achieving significantly higher overall hardware efficiency.

\begin{listing}[ht]
\begin{minted}[
  fontsize=\small,
  breaklines,
  breakanywhere,
  baselinestretch=0.95,
  bgcolor=CodeBg,  % <--- background color here
  fontfamily=zi4  % uses Inconsolata if available
]{python}
@register_dispatcher("async_batch_bounded")
async def async_batch_bounded(cfg, trajs, init_fn, run_fn, eval_fn):
    sem = asyncio.Semaphore(cfg.get("max_parallel_agents", 8))
    async def one(i, j):
        async with sem:
            tr = trajs[i][j]
            await getattr(tr, init_fn)()
            await getattr(tr, run_fn)()
            await getattr(tr, eval_fn)()
    await asyncio.gather(*[asyncio.create_task(one(i, j)) for i in trajs for j in trajs[i]])

\end{minted}
\caption{Simplified Implementation for Async Batch (Bounded).}
\label{lst:dispatcher_example}
\end{listing}

An example implementation of \emph{Async Batch (Bounded)} is provided in \cref{lst:dispatcher_example}, demonstrating that new dispatching strategies can be added with minimal code under the unified interface. This design makes it easy to implement customized parallelization strategies for different workload characteristics, for example, priority-based scheduling that prioritizes trajectories with higher evaluation cost (e.g., more test cases), preventing them from becoming long-tail stragglers and reducing the overall makespan.

\subsection{Connecting to training backends.}\label{sec:backend}

We adopt a \textit{transition-based design} that records each LLM invocation as an individual transition containing the input tokens, output tokens, and their corresponding log probabilities (if provided by the inference endpoint). These transitions form the foundation for constructing the final training data and enable several key capabilities for robust RL training.
\emph{First, it can help mitigate consistency between inference and training.} By explicitly recording log probabilities, we can apply recent techniques such as Flash-RL~\citep{liu2025flashrl} to correct for inference–training engine mismatch~\citep{yao2025offpolicy}.
\emph{Second, it guarantees token-level fidelity.} The transition format naturally supports token-in/token-out processing, eliminating off-policy drift caused by re-tokenization or text postprocessing~\citep{tito}.
\emph{Third, it enables algorithmic flexibility.} Transition-level data enables a wider range of RL algorithms beyond mask-based concatenation approaches, which are often inefficient and rigid. For example, we can dynamically summarize prior context (see examples in \cref{sec:resum_agent}), insert structured prompts, or even modify the agent’s roles across turns, thereby enabling more diverse and flexible agent training methods. The recording mechanism is implemented via a lightweight decorator that transparently captures model inputs and outputs during generation, creating transition objects without requiring any modification to agent logic:
\begin{minted}[
  fontsize=\small,
  breaklines,
  breakanywhere,
  baselinestretch=0.95,
  bgcolor=CodeBg,  % <--- background color here
  fontfamily=zi4  % uses Inconsolata if available
]{python}
@record_transition
async def _generate(self, input_ids, sampling_params):
    return await self.backend.generate(input_ids, sampling_params)
\end{minted}
Additionally, for computational efficiency, we dynamically pack transitions that share the same prefix into a single training sample using a standard masking-based method. As a result, for trajectories without any context modification during execution, the system naturally falls back to the traditional concatenation-and-mask approach.

After execution, the runner aggregates the training samples into a unified intermediate format, as shown in \cref{lst:postprocess}.
\frameworkname transparently converts this format into the required input structure for different backends for advantage estimation and policy updates.
As a result, switching from local evaluation to distributed training only requires adjusting configuration rather than modifying agent or task code, preserving a clean separation of concerns while enabling backend specialization.

\begin{listing}[h]
\begin{minted}[
  fontsize=\small,
  breaklines,
  breakanywhere,
  baselinestretch=0.95,
  bgcolor=CodeBg,  % <--- background color here
  fontfamily=zi4  % uses Inconsolata if available
]{python}
def post_process(...):
    return {
        "prompt_token_ids": ...,
        "response_ids": ...,
        "logprobs": ...,
        "loss_masks": ...,
        "traj_rewards": ...,
        "traj_idx": ...,
        "rollout_metrics": ...,
    }
\end{minted}
\caption{Backend-agnostic training data returned by \texttt{post\_process}.}
\label{lst:postprocess}
\end{listing}

By abstracting away the training backend, researchers can compare alternative approaches on identical tasks without confounds from differing implementations, accelerating algorithmic progress. Equally, the abstraction preserves access to backend-specific strengths such as advanced model support, novel training algorithms, specialized hardware support (e.g., GPU/TPU/AMD accelerators), and API-based execution for very large models, so practitioners can exploit these capabilities through configuration without changing the code.

\subsection{Error Handling} 
Agent execution in realistic environments inevitably encounters various failures: LLM generation issues, malformed tool calls, execution timeouts, and resource exhaustion. While our tool‑centric design already localizes many environment‑specific errors to the tool level, \frameworkname additionally handles two classes of failures \emph{in the agent loop}: \emph{terminal} conditions (e.g., context‑window exceeded, max‑iterations reached), which immediately stop the episode, and \emph{recoverable} conditions (e.g., parse failures, incorrect parameters), for which the framework injects corrective feedback into the conversation history to enable self‑correction on the next step, so that agents can recover from these failures, improving overall success rates.

\section{SWE Agent}
We will now present the full training recipe for \model, implemented using \frameworkname.
\subsection{Background}

Recent work has rapidly advanced the use of large language models (LLMs) for automating
software engineering (SWE) tasks \citep{jimenez2024swebench,yang2024swe,yang2025kimi,pan2024training}.
\textsc{SWE‑Bench} \citep{jimenez2024swebench} has emerged as a central testbed: given a real GitHub repository and a bug report, a system
must produce a patch and is judged by whether the project’s unit tests pass after applying
that patch.

\begin{wrapfigure}{r}{0.48\textwidth}
  \centering
  \includegraphics[width=\linewidth]{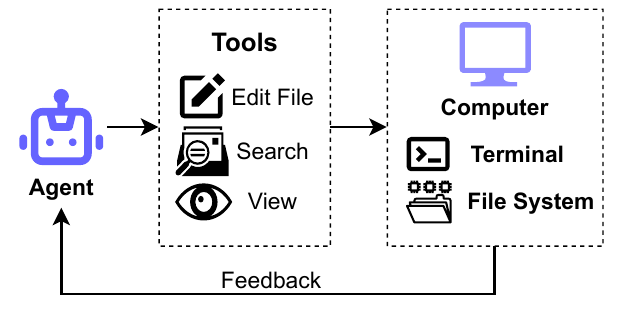}
  \caption{Illustration of an SWE Agent.}
  \label{fig:swe_agent}
\end{wrapfigure}

Current solutions on SWE tasks cluster into two families. \emph{Agent‑based} systems~\citep{wang2024openhands,yang2024swe} place the model in an executable environment with a set of tools (e.g., shell, search, and file viewer/editor) and let it iteratively inspect, edit, build, and test until completion, as illustrated in~\cref{fig:swe_agent}. This mirrors developer workflows and supports incremental diagnosis and repair. \emph{Workflow‑based} systems such as Agentless~\citep{xia2024agentless}, by contrast, prescribe a fixed pipeline (e.g., localize $\rightarrow$ repair $\rightarrow$ test), reducing the problem to a chain of verifiable single‑turn sub-tasks. The former favors flexibility and fidelity to practice; the latter simplifies orchestration and evaluation at the cost of interactivity and generalizability.

\frameworkname is designed to facilitate end‑to‑end RL training of long‑horizon, multi‑turn, multi‑tool agents across diverse agentic tasks. We posit that the tool‑use competence acquired in SWE, where agents must iteratively inspect, edit, and test real codebases, transfers to other domains that similarly require stateful tool interaction. Consequently, in this work, we focus more on training the full interactive loop in an end-to-end manner.

\subsection{Training Recipe}\label{sec:swe_recipe}
\paragraph{Bootstrapping training with better tools.}
Recent studies have shown that agents often struggle with error localization, a key bottleneck in software-enginerring tasks~\citep{chen2025locagent,zan2025multi}.
We observe that this failure largely arises from the agent’s inability to use commands to effectively navigate the codebase. Specifically, agents tend to over-rely on viewing files directly instead of leveraging search utilities (e.g., \texttt{grep}, \texttt{find}), and often fail to refine their queries with informative patterns or precise keywords, leading to repeated failures to locate relevant context. As a result, the agent frequently falls back to using the \texttt{view} tool to look through files chunk by chunk, or retrieves large amounts of irrelevant content, both of which consume excessive context and reduce overall efficiency.


Based on these observations, we implement an AST-based search tool that provides powerful retrieval capabilities supporting both fuzzy matching and structural pattern search, inspired by LocAgent~\citep{chen2025locagent}. To further encourage the search tool usage, we append contextual hints at the end of each search result (e.g., suggesting which specific terms or patterns to search next) to guide the agent toward more precise queries. 
We find that such tool enhancements are critical for efficient RL training. Without them, end-to-end RL becomes extremely challenging due to the sparse and delayed rewards inherent in long-horizon tasks. In particular, minimal-tool configurations, such as the \textbf{bash-only} setup used in Mini-SWE-Agent~\citep{minisweagent}, exhibit very high non-resolved rates (e.g., 50/64) on the R2E-Gym data, making the learning process challenging and inefficient. In contrast, our optimized, tool-guided setup significantly accelerates convergence, achieving comparable or even higher performance than prior models of similar size within only \textbf{125} steps of RL training (\cref{fig:swe_reward}). For evaluation, to assess generalization, we ablate the search tool. Remarkably, the model exhibits emergent, internally realized search behavior and attains comparable performance without explicit access to a dedicated search tool. We also observe in the training that as training progresses, there is a steady increase in the average number of search invocations per trajectory, indicating that the agent gradually learns to rely more on search for efficient code localization, as shown in \cref{fig:swe_reward}.

\paragraph{RL algorithms and hyper-parameters.}
For stable training, we adopt a \emph{fully on-policy} setup where the training batch size equals the mini-batch size (both set to 64), and the number of rollouts per task is 8. Trajectories that terminate due to external constraints, such as the maximum context length (32K tokens), step limit (50 turns), are masked out during gradient updates to prevent the model from being biased against trajectories with more actions or reasoning steps. Importantly, this masking does not modify the reward or advantage estimation; it only excludes those samples from the gradient computation. Following prior works~\citep{deepswe2025,liu2025understanding}, we apply leave-one-out advantage estimation and remove both standard deviation and length normalization in advantage computation. We disable KL and entropy loss and train with a learning rate of $1e-6$. The Qwen3 chat template is modified to preserve the ``thinking'' from previous turns, maintaining reasoning continuity across steps.

\paragraph{Hints for agent to recover and proceed.}
Multi-turn agent reinforcement learning often suffers from agents getting stuck in repetitive or unproductive behaviors. 
Prior work, such as SWE-Gym~\citep{pan2024training} and Kimi-Dev~\citep{yang2025kimi} observes that starting from a generic pre-trained model can lead to brittle behaviors: failing to invoke tools, generating no or incorrect function calls, or looping over similar actions without progress. 
Agents may also lose track of the task context or remaining steps, and tend to stop solving the problem prematurely without reaching the maximum turn or context limit. 
During training, \emph{hints} are introduced as structured cues that help agents recover from failed actions and re-enter valid trajectories. 
Such hints include suggestions about potential actions to take upon tool execution failure, notifications that the remaining step budget or context window is about to be exceeded, corrections for invalid or incomplete function calls, or prompts to re-examine failed edits. These hints substantially improve trajectory quality, stabilize rollout collection, and increase the proportion of successful trajectories used for policy optimization.

\subsection{Evaluations}\label{sec:swe_eval}
To demonstrate the effectiveness of our RL training, we evaluate \model across 4 representative agentic benchmarks. These include \textbf{SWE-Bench Verified}~\citep{jimenez2024swebench} as the in-domain evaluation, and three out-of-domain benchmarks: \textbf{Terminal-Bench}~\citep{tbench_2025}, \textbf{WebArena}~\citep{zhou2023webarena}, and \textbf{BrowseComp-Plus}~\citep{chen2025browsecomp}, which assess the model’s generalization to diverse interactive environments involving command-line reasoning, web navigation, and search-integrated reasoning.

\paragraph{SWE-Bench Verified} We evaluate \model and several baseline models on SWE-Bench verified using a simple ReAct~\citep{yao2022react} scaffold, and the models are provided with only a bash tool and a file editor tool\footnote{\url{https://github.com/All-Hands-AI/openhands-aci/blob/main/openhands_aci/editor/editor.py}}.
The maximum context length is set to 40K (the default for Qwen3-32B), and the maximum number of steps to 100.
The main results are reported in~\cref{tab:swe_results}. We highlight that, \model achieves state-of-the-art performance across open-recipe models at its scale, on SWE-Bench Verified, under an end-to-end interactive evaluation setting, demonstrating strong agentic tool-use ability, while incurring substantially lower training cost and without relying on distillation from a stronger teacher model.

Among the baseline models, SWE-agent-LM-32B~\citep{yang2025swe} is fine-tuned from Qwen2.5-Coder-32B-Instruct on 5,016 SWE-smith~\citep{yang2025swe} trajectories generated by Claude Sonnet 3.7. DeepSWE~\citep{deepswe2025}, on the other hand, is trained purely with reinforcement learning on the same 4.5K R2E-Gym dataset using the default R2E-Gym scaffold. Compared to DeepSWE, we enhance the training recipe with tool-guided RL, which significantly improves training efficiency. As shown in~\cref{fig:swe_reward}, the non-resolved rate of \model remains consistently lower and decreases faster than that of DeepSWE. Combined with the system-level optimizations introduced in~\cref{sec:design}, our approach achieves 50\% lower training cost while delivering superior performance. Models such as Kimi-dev \citep{yang2025kimi} and SWE-Swiss \citep{SWESwiss2025}, which are trained under the Agentless \citep{xia2024agentless} scaffold, struggle to follow tool-call instructions in the ReAct setup, leading to substantially lower scores than those reported in their original works, and we therefore only include their reported results in the table for reference.

\begin{table*}[ht]
  \centering
  \caption{Pass@1 performance on SWE-Bench Verified for reference and open-source SWE models. ``--'' denotes unavailable values. ``$\times$'' denotes omitted results. Reported scores are taken from the corresponding papers/blogs; because different works use distinct scaffolds and evaluation setups, the numbers may not be strictly comparable. \emph{Simple ReAct} denotes a minimal ReAct agent loop that exposes only a bash tool and a file‑editing tool and the model generates only one patch per instance.}
  \small
  \renewcommand{\arraystretch}{1.22}
  \resizebox{0.95\textwidth}{!}{%
    \begin{tabular}{lllccc}
      \toprule
      \multirow{2}{*}{\textbf{Model}} &
      \multirow{2}{*}{\textbf{Model Size}} &
      \multirow{2}{*}{\textbf{Recipe}} &
      \multicolumn{2}{c}{\textbf{SWE-Bench Verified}} &
      \multirow{2}{*}{\textbf{Cost (H100 Hours)}} \\
      \cmidrule(lr){4-5}
      & & & Simple ReAct & Reported & \\
      \midrule
      \rowcolor{gray!12}\multicolumn{6}{l}{\textit{Reference}} \\
      Qwen3-32B           & 32B & --                & 24.4 & --               & -- \\
      Qwen3-Coder-30B     & 30B & --                & 45.0 & --               & -- \\
      \midrule
      \rowcolor{gray!12}\multicolumn{6}{l}{\textit{Open-source models $<$100B}} \\
      SWE-agent-LM-32B    & 32B & 3.7 Sonnet Distill& 38   & 40.2 & -- \\
      SWE-Swiss           & 32B & R1 Distill + RL   & $\times$   & 45.0 & -- \\
      Kimi-dev            & 72B & R1 Distill + RL   & $\times$  & 48.6 & -- \\
      DeepSWE             & 32B & RL                & 36.4 & 42.2 & 9180 \\
      \rowcolor{blue!8}
      SA-SWE-32B          & 32B & RL                & \textbf{39.4} & --               & 4601 \\
      \bottomrule
    \end{tabular}%
  }
  \label{tab:swe_results}
\end{table*}

\paragraph{Terminal-Bench} Terminal-Bench (version 0.1.1) is a command-line interaction benchmark designed to evaluate agents on system-level tasks that require sequential command execution, environment manipulation, and output verification. It contains 80 tasks spanning categories such as software engineering, system administration, and security, measuring an agent's ability to navigate file systems, execute shell commands, and complete multi-step technical workflows. In our evaluation, we use the OpenHands agent\footnote{\url{https://github.com/laude-institute/terminal-bench/tree/main/terminal_bench/agents/installed_agents/openhands}} to compare the baseline Qwen3-32B model against \model. 
\paragraph{WebArena} WebArena is a web-based interaction benchmark designed to evaluate agents on realistic website navigation and task completion across multiple domains. It contains 812 tasks spanning e-commerce, social forums, collaborative software, and content management systems, measuring an agent's ability to interpret web interfaces, execute multi-step workflows, and achieve specified goals through browser interactions. The benchmark uses fully functional websites with dynamic content to simulate real-world web automation scenarios. In our evaluation, we compare the baseline Qwen3-32B model against \model.

\begin{wraptable}{r}{0.55\textwidth}
  \centering
  \caption{\model's performance on other tasks.}
  \small
  \renewcommand{\arraystretch}{1.22}
  \resizebox{0.53\textwidth}{!}{%
    \begin{tabular}{lcccc}
      \toprule
      \multirow{2}{*}{\textbf{Model}} 
        & \multirow{2}{*}{\textbf{Terminal Bench}} 
        & \multicolumn{2}{c}{\textbf{BrowseComp-Plus}} 
        & \multirow{2}{*}{\textbf{WebArena}} \\
      \cmidrule(lr){3-4}
        &  & Acc. & Avg. Turn &  \\
      \midrule
      \rowcolor{gray!12}\multicolumn{5}{l}{\textit{Reference}} \\
      Qwen3-32B  & 13.75 & 18.1 & 3.68 & 15.8  \\
      \midrule
      \rowcolor{blue!8}
      SA-SWE-32B & \textbf{16.25} & \textbf{19.4} & \textbf{4.6} & \textbf{17.0} \\
      \bottomrule
    \end{tabular}%
   }
\label{tab:ood_results}
\end{wraptable}

\paragraph{BrowseComp-Plus} BrowseComp-Plus is a search-oriented benchmark designed to evaluate agents on iterative search, reasoning, and planning tasks. It uses a fixed, human-verified corpus containing both supporting and hard-negative documents to ensure fair and reproducible evaluation. The benchmark consists of 830 questions and measures an agent’s ability to plan multi-step searches, reason over retrieved evidence, and generate accurate final answers. In our evaluation, we use Qwen3-Embedding-8B~\citep{zhang2025qwen3} as the retriever and configure the search tool to return top-5 search results.

As demonstrated in~\cref{tab:ood_results}, \model demonstrates improved performance over the base model across all of the benchmarks. During SWE agent training (refer to~\cref{fig:swe_reward}), we observe that the average number of search calls increases from 3 to 4, and the average number of turns grows from 18 to 25, indicating that the agent learns to rely more on external search and iterative reasoning as training progresses. Remarkably, on BrowseComp-Plus, which requires extensive search and reasoning across multiple documents, our model, although trained only on the SWE task, exhibits a similar trend by making more search calls than the base model.

\section{Other Agents}\label{sec:other_agents}
In this section, we provide additional examples implemented in~\frameworkname, including deep research agents, a special agent that manages its own memory, and computer use agents.

\subsection{Deep Research}
\label{sec:research_agent}

Deep Research agents~\citep{deepresearch,gao2025beyond,li2025webthinker,li2025websailor} are designed to tackle comprehensive question-answering tasks where the required knowledge is not contained within the model parameters. These agents typically rely on web-scale search (e.g., Google Search) and browser-based interaction to gather, reason over, and synthesize external information.


In \frameworkname, we support both searching and web-browsing tools under the unified tool-centric interface. Specifically, we adopt the ReAct framework, using SerperAPI for search and Jina Reader API~\citep{jina} for web-page retrieval. To improve efficiency, we introduce caching to avoid repeated fetches and a two-stage summarization pipeline that splits long pages into chunks, summarizes each, and merges the results while preserving references. These optimizations significantly accelerate rollouts and reduce repeated failures during multi-turn research tasks. 


\paragraph{Experiment setup.}
Recent work~\citep{Polaris2025,cheng2025revisiting} highlights the importance of balancing both \emph{difficulty} and \emph{diversity} in reinforcement learning data. To this end, we sample 50K problems from~\citep{fan2025megascience} and perform offline difficulty estimation using Qwen3-8B in thinking mode, generating four rollouts per example. Each problem is categorized by the number of successful rollouts: 0/4 (Impossible), 1/4 (Hard), 2/4 (Medium), 3/4 (Easy), and 4/4 (Perfect). Following the mirrored difficulty distribution proposed in Polaris, we construct a dataset composed of 25\% Impossible, 30\% Hard, 30\% Medium, and 15\% Easy examples, further balanced across STEM disciplines (computer science, biology, physics, and economics) to ensure broad coverage and representational diversity. All experiments are trained using the SkyRL-train~\citep{griggs2025evolving} backend.


\begin{wrapfigure}{r}{0.45\textwidth}
  \centering
  \includegraphics[width=\linewidth]{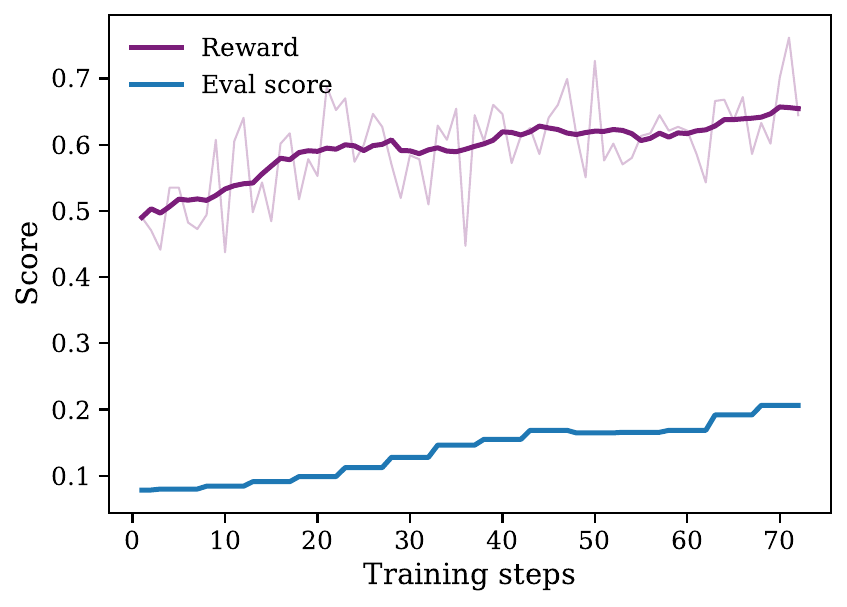}
  \caption{Training Curve for Deep Research Agent on SkyRL-train backend.}
  \label{fig:research_curve}
\end{wrapfigure}


We train Qwen3-8B with GRPO~\cite{shao2024deepseekmath} using the general verifier from~\citep{ma2025general}, a global batch size of 64 with 8 rollouts, a mini-batch size of 64, and a learning rate of $5\times10^{-6}$. The training uses Qwen3-235B in non-reasoning mode as the summarizer. Reward and validation curves during training are shown in~\cref{fig:research_curve}. For evaluation, we use LLM-as-a-Judge on the \textsc{HLE-500} benchmark~\citep{li2025webthinker}. Since the general verifier~\citep{ma2025general} tends to be more lenient, we additionally use gpt-oss-20b as the primary judge. Under the general verifier, scores improve from 12.6\% to 18.8\%, and under gpt-oss-20b, from 9.2\% to 10.2\% / 11.0\%.

\textbf{Challenges.} We encountered two challenges in deep-research training. \emph{First}, heavy tools introduce significant compute overhead and, if under-provisioned, harm both throughput and stability. Web-scraping pipelines ($\sim$10k tokens/request) paired with large summarizers (e.g., Qwen-3-32B) slowed training drastically: serving Qwen-3-32B on a single H200 took 2,101.6 s per iteration, whereas switching to a Qwen API reduced this to 592.1 s ($\sim$3.5$\times$ faster). Our final stage used four GH200s with a data-parallel router. Insufficient serving caused timeouts that contaminated rollouts (e.g., a dip around iteration 30) and required multiple steps (30–33) to recover, motivating both adequate tool capacity and robust handling (e.g., masking abnormal trajectories) to protect the reward signal.
\emph{Second}, online search can leak benchmark answers. During evaluation on GPQA-Diamond, models sometimes retrieved solutions directly from public pages (e.g., Hugging Face) instead of reasoning, inflating scores via shortcut retrieval. We therefore enforce domain blocks (e.g., Hugging Face, GitHub, GitLab, Chegg) to limit leakage and encourage authentic, tool-augmented reasoning.

\subsection{Memory Agent}
\label{sec:resum_agent}

\begin{wrapfigure}[22]{r}{0.46\textwidth} 
  \vspace{-4pt}
  \centering

  \begin{subfigure}[t]{\linewidth}
    \centering
    \includegraphics[width=\linewidth]{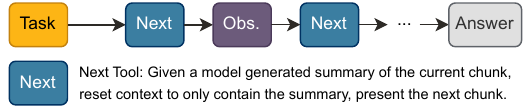}
    \caption{\textbf{MemAgent Scaffold in \frameworkname.} 
    The agent processes document chunks using the \texttt{Next} tool, which summarizes the current chunk and appends the summary to the next chunk as input for the following turn.}
    \label{fig:memagent_scaffold}
  \end{subfigure}

  \vspace{4pt}

  \begin{subfigure}[t]{\linewidth}
    \centering
    \includegraphics[width=0.8\linewidth]{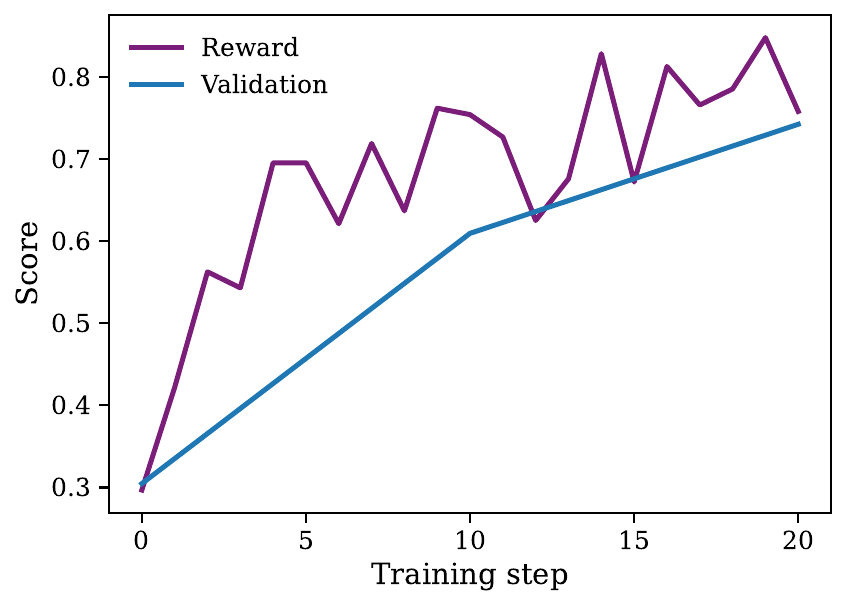}
    \caption{\textbf{Training Reward and Validation Score.} 
    Results on RULER-HotpotQA for the memory agent trained from the Tinker backend.}
    \label{fig:memagent_curve}
  \end{subfigure}

  \vspace{-6pt}
  \caption{MemAgent Results.}
  \label{fig:memagent_combined}
\end{wrapfigure}

The fixed context length remains a fundamental limitation for contemporary LLMs. Recent work such as MemAgent~\citep{yu2025memagent} proposes a recursive long-text processing mechanism, where the model incrementally summarizes content into a fixed-size memory buffer. We implemented the MemAgent scaffold by introducing a \texttt{Next} tool, as illustrated in~\cref{fig:memagent_scaffold}.

\textbf{Experimental Setup.} We evaluate on the RULER-HotpotQA dataset~\citep{hsieh2024ruler}, which provides a golden supporting paragraph accompanied by several distractor paragraphs. 
Following the setup of MemAgent, the model is presented with chunks of text up to 4k tokens each. 
During training, it processes sequences of up to 28k tokens, and during evaluation, up to 112k tokens. We use GPT-5-nano as the verifier for answer correctness.
We train the model using the Tinker backend from Qwen3-8B (non-thinking mode), using a LoRA rank of 128, batch size 32, 8 rollouts, and a maximum of 8K tokens per turn. Training curves are shown in~\cref{fig:memagent_curve}. For reference, MemAgent reports a 79.69\% accuracy when training Qwen2.5-7B-Instruct under the same dataset using a batch size of 128 and group size 16, evaluated with an exact-match verifier. 



\subsection{Computer Use Agent}\label{sec:computer_agent}
Computer Use agents aim to complete tasks via interacting with an actual computer interface.
We implement Computer Use agent training using the OSWorld environment~\citep{OSWorld} where the model can interact with a virtual computer through the PyAutoGUI APIs and get environment feedback through accessibility trees generated from the virtual desktop. We integrate the OSWorld environment as a tool in \frameworkname, as shown in~\cref{lst:osworld}, enabling agents to perform GUI interactions such as mouse clicks, keystrokes, scrolling, and window manipulation directly through generated Python codes with PyAutoGUI APIs.

\begin{wrapfigure}[11]{l}{0.42\textwidth}
  \vspace{-10pt}
  \centering
  \includegraphics[width=0.8\linewidth]{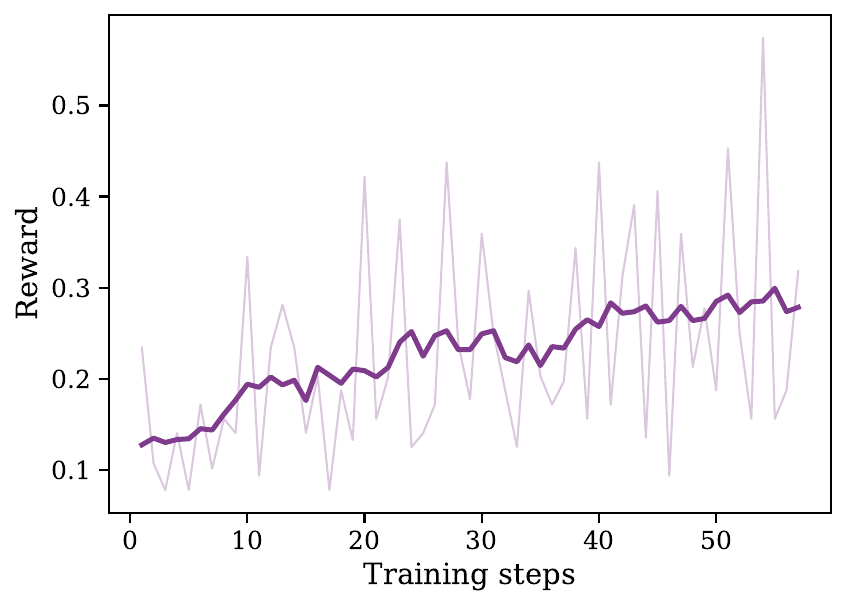}
  \vspace{-10pt}
  \caption{Training Curve for Computer Use Agent on VeRL backend.}
  \label{fig:osworld_curve}
\end{wrapfigure}

\paragraph{Experiement Setup.} Due to the lack of data, we mirror the setup in ARPO~\citep{lu2025arpo}, which chooses a subset from the benchmark to be the training data. Similar to the data filtering process for the Deep Research task (\cref{sec:research_agent}), we filter the OSWorld benchmark and collect a subset of 32 tasks with Hard, Medium, and Easy difficulties.

We train Qwen3-8B with GRPO and implement the rollout process using the \textbf{Async Batch (Bounded)} dispatcher with fixed 32 virtual desktop environments. We scale the OSWorld environments by launching its virtual-machine as Ray~\citep{moritz2018ray} remote tasks, disaggregating CPU-bound feedback loops from GPU-bound model generation. The training batch size is set to 8 with 8 rollouts for each task. The evaluation is directly performed by the environment’s native evaluation function, with training curves shown in~\cref{fig:osworld_curve}.

\paragraph{Observations.} We observe that for OSWorld tasks, while the training reward steadily improves, the validation accuracy shows little to no gain. This suggests that the tasks are inherently difficult for Qwen3-8B, and the learned policy struggles to generalize beyond the training environments. Similar trends have also been reported in~\citep{lu2025arpo}.

\section*{Acknowledgment}
This work is supported by the kind compute support from Anyscale, Amazon, Lambda, Thinking Machine Lab, and 
gifts from AMD, Google, Mayfield, Mithril, Laude Institute, Accenture, Broadcom, Cisco, IBM, Intel, Intesa Sanpaolo, Lightspeed, Mibura, Microsoft, NVIDIA, Samsung SDS, and SAP. We also thank John Yang, Naman Jain, Xingyao Wang, Jiayi Pan, Chuan Li and Eric Tu for helpful discussion and feedback.

\bibliography{ref}
\bibliographystyle{iclr2026_conference}

\clearpage

\end{document}